\documentclass[conference]{IEEEtran}
\IEEEoverridecommandlockouts
\usepackage{amsmath,amssymb,amsfonts}
\usepackage{algorithmic}
\usepackage{cite}
\usepackage{mathtools}
\usepackage[caption=false]{subfig}

\usepackage{booktabs}
\usepackage{graphicx}
\usepackage{amsthm}

\newtheorem*{examplebox*}{Example}

\usepackage{tikz}
\usetikzlibrary{automata}
\usetikzlibrary{shapes.geometric, arrows.meta, positioning, fit, backgrounds}
\usetikzlibrary{calc}

\begin{document}

\title{Explainable Scene Understanding with Qualitative Representations and Graph Neural Networks\thanks{This work is funded by the European Commission through the AI4CCAM project (Trustworthy AI for Connected, Cooperative Automated Mobility) under grant agreement No 101076911, and by the Research Council of Norway through the AutoCSP project, grant number 324674.}}

\author{\IEEEauthorblockN{Nassim Belmecheri}
\IEEEauthorblockA{\textit{VIAS Department} \\
\textit{Simula Research Laboratory}\\
Oslo, Norway \\
nassim@simula.no}
\and
\IEEEauthorblockN{Arnaud Gotlieb}
\IEEEauthorblockA{\textit{VIAS Department} \\
\textit{Simula Research Laboratory}\\
Oslo, Norway \\
arnaud@simula.no}
\and
\IEEEauthorblockN{Nadjib Lazaar}
\IEEEauthorblockA{\textit{LISN} \\
\textit{Université Paris-Saclay}\\
Saclay, France \\
lazaar@lisn.fr}
\and
\IEEEauthorblockN{Helge Spieker}
\IEEEauthorblockA{\textit{VIAS Department} \\
\textit{Simula Research Laboratory}\\
Oslo, Norway \\
helge@simula.no}
}

\maketitle

\begin{abstract}
This paper investigates the integration of graph neural networks (GNNs) with Qualitative Explainable Graphs (QXGs) for scene understanding in automated driving. 
Scene understanding  is the basis for any further reactive or proactive decision-making. Scene understanding and related reasoning is inherently an explanation task: why is another traffic participant doing something, what or who caused their actions?
While previous work demonstrated QXGs' effectiveness using shallow machine learning models, these approaches were limited to analysing single relation chains between object pairs, disregarding the broader scene context. We propose a novel GNN architecture that processes entire graph structures to identify relevant objects in traffic scenes. We evaluate our method on the nuScenes dataset enriched with DriveLM's human-annotated relevance labels. Experimental results show that our GNN-based approach achieves superior performance compared to baseline methods. The model effectively handles the inherent class imbalance in relevant object identification tasks while considering the complete spatial-temporal relationships between all objects in the scene. Our work demonstrates the potential of combining qualitative representations with deep learning approaches for explainable scene understanding in autonomous driving systems.
\end{abstract}


\section{Introduction}

Scene understanding is a central task in automated driving as the basis for informed decision-making.
Scene understanding takes the perception of the environment and derives information about other traffic participants, such as other vehicles, vulnerable road users (VRUs), or infrastructure, such as traffic lights or road signs~\cite{yang_scene_2019,muhammad_vision-based_2022,muhammad_deep_2021}.

Recently, the field has seen increasing interest in explainable representations for traffic scenes. One advancement in this direction is the \emph{Qualitative Explainable Graph} (QXG), introduced by Belmecheri et al. \cite{belmecheri2023acquiring,belmecheri_2024_tra,belmecheri2024trustworthy}. QXGs are a symbolic graph structure that represent traffic scenes through qualitative spatio-temporal relationships.
This representation been shown to be effective for explanations in automated driving settings, such as relevant object identification~\cite{belmecheri2024roi} and action explanations~\cite{belmecheri2024trustworthy}.

However, current approaches to processing QXGs rely primarily on shallow machine learning models, such as random forests. While these models demonstrate good performance and support interpretation of the explanation process, they have a significant limitation: they can only process single relation chains between pairs of objects, independent of other objects in the scene. This restricted view fails to capture the rich contextual information present in traffic scenes, where the relevance or behaviour of objects often depends on their relationships with multiple other participants.

However, current approaches to processing QXGs rely primarily on shallow machine learning models, for example, random forests.
While these models show good performance and support the interpretation of the explanation process, they show one major limitation: they do not consider the entire QXG, but only a single relation chain between two objects, independent of the other objects in the scene.
This restricted view fails to capture the rich contextual information present in traffic scenes, where the relevance or behaviour of objects often depends on their relationships with multiple other participants.

In this paper, we explore the combination of QXGs with graph neural networks (GNNs), a class of deep learning models specifically designed to handle graph-structured data. Our approach enables processing of complete graph structures, allowing the model to consider all object relationships simultaneously when making predictions. Our contributions are threefold: 
(1) We present a novel GNN architecture designed to process QXGs while maintaining their explainable properties. 
(2) We introduce a specialized training approach that handles the inherent class imbalance in relevant object identification. 
(3) We demonstrate improved performance over traditional shallow learning methods through comprehensive experiments on real-world driving data.

Our work bridges the gap between qualitative scene representations and advanced deep learning techniques, showing that it's possible to leverage the power of neural networks from an inherently symbolic and explainable scene representation, such as the QXG.

\section{Background}

\subsection{Qualitative Explainable Graphs}\label{sec:qxg}

\begin{figure}[t]
    \centering
    \includegraphics[width=\columnwidth]{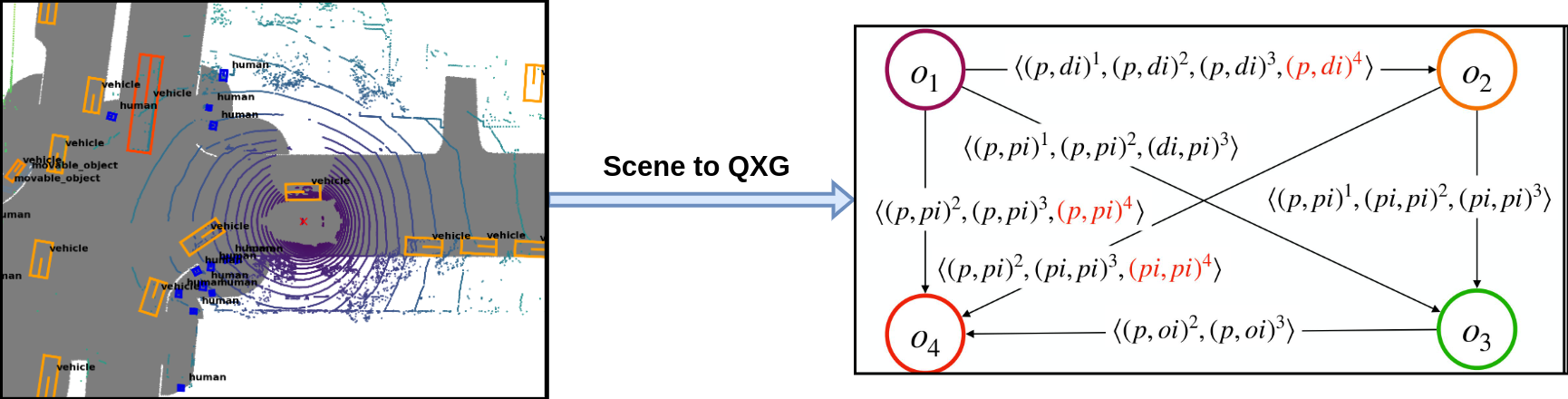}
    \caption{Illustration of the QXG built from a scene}
    \label{fig:qxgbuilder}
\end{figure}

The Qualitative Explainable Graph (QXG) is a format for scene representation that describes the qualitative spatial-temporal relations among objects within a scene. 
The QXG was first introduced in \cite{belmecheri2023acquiring}, and was expanded into a more comprehensive representation with multiple qualitative calculi in \cite{belmecheri2024trustworthy}. 
The graph that represents a scene, denoted as $S$, consists of individual nodes for each object in $\mathcal{O}$. 
Edges $\mathcal{V}$ are drawn between objects that appear together in at least one frame $f$. Each edge is labelled with spatial relations, that are computed per frame between pairs of objects $(o_i,o_j) \in \mathcal{O}$. 
Formally, let $\mathcal{A}=\{A_1,\ldots, A_n\}$ be a set of algebras, where each algebra $A_i$ is characterized by a set of relations $R_i=\{r_1,\ldots,r_m\}$ that describe a specific spatial relation $r_i$ between pairs of objects $(o_i,o_j) \in \mathcal{O}$ at a frame $f_k$. 
For a given scene $s_i \in S$ there are multiple frames, thus for each frame $f_k$ we have a relation $r_k$ per algebra $A_i$, which gives the spatio-temporal property to the QXG. 
Figure \ref{fig:qxgbuilder} represents the QXG built from a scene of 3 frames, only the 3rd frame is highlighted in the figure, but the QXG contains the spatio-temporal interactions of the 3 frames.

For the purpose of this paper, we describe relationships between objects using a combination of the three calculi mentioned earlier to capture the necessary spatial information, namely \emph{Qualitative Distance Calculi}~\cite{renz_qualitative_2007} for distance, \emph{Qualitative Trajectory Calculi}~\cite{dylla_survey_2015} for trajectory dynamics, and \emph{Rectangle Algebra}~\cite{renz_qualitative_2007} for relative positioning. Consider the following example:

\begin{examplebox*}
Given two detected objects $o_1$ (car) and $o_2$ (pedestrian) in a frame $f_k$, we have their corresponding bounding boxes $bbox_1$, $bbox_2$ and if we consider \emph{Qualitative Distance Calculi} and a spatial algebra then the possible relations would be $\emph{\{very close, close, far, very\ far\}}$, these can be computed based on the Euclidean distance computed from the centroids of $bbox_1$ and $bbox_2$. 
If $distance(bbox_1,bbox_2)$ is less, greater, or between certain thresholds $\theta$, then the distance relation is set accordingly.
\end{examplebox*}

However, it is important to note that the formulation and application of the QXG are generally not dependent on the specific calculi chosen, provided they are expressive enough to cover at least the relative positioning and distance of objects. Depending on the use case, additional calculi might enhance the representation~\cite{dylla_survey_2015}.

\subsection{Scene Understanding \& Explanations}
Scene graphs have been utilized for various applications, including indoor \cite{yang2017support}, video \cite{gay2019visual, mavroudi2020representation}, and 3D scene understanding \cite{wald2020learning}, as well as for providing scene explanations \cite{shi2019explainable}. 
In automated driving, scene understanding must include vehicles, vulnerable road users, and stationary elements like traffic lights and barriers, along with their movements. Scene graphs have been applied for overall scene understanding \cite{guo2012semantic1, greve2023collaborative}, lane estimation \cite{zurn2021lane}, and action prediction \cite{kochakarn2023explainable}.

Let \(\mathcal{S}\) represent a scene containing a set of objects \(\mathcal{O} = \{o_1, o_2, \ldots, o_n\}\). Each object \(o_i\) is detected using an object detection algorithm and is represented by a 3D bounding box \(b_i\).

Let \(T\) be the set of time frames in the scene under consideration.
Let \(R\) be the set of possible qualitative spatial relations.

Scene understanding involves identifying and analysing the spatio-temporal relationships between these objects.

A scene \(\mathcal{S}\) can be represented by a set of tuples \(\{(o_i, o_j, r_t) \mid o_i, o_j \in \mathcal{O}, r \in R, t \in T\}\), where:
\begin{itemize}
    \item \(o_i\) and \(o_j\) are objects in the scene.
    \item \(r\) is a qualitative spatial relation between \(o_i\) and \(o_j\) at time \(t\) .
\end{itemize}

An explanation \(X\) is defined as a relation chain of \(R\) within a time interval \(T\) between a pair of objects such that either one of the objects is relevant or causes an event in the scene.

Let \(A^*\) be the set of actions or events of interest. \(\mathcal{E}\) denotes the set of explanations corresponding to these actions or events. Let \(\mathcal{C} \subseteq \mathcal{O}\) be the set of objects that are relevant or cause an event.

An explanation \(X \in \mathcal{E}\) for an action \(a \in A^*\) can be defined as a set of relation tuples \(\{(o_i, o_j, r_t) \mid (o_i \in \mathcal{C} \vee o_j \in \mathcal{C}), r \in R, t \in T\}\).

\section{Related Work}

Scene understanding involves collecting, organizing, and analysing spatial and temporal information about various objects. One effective method is modelling the relations between individual objects as a scene graph \cite{scene_graph_survey}.

Scene graphs have been used in various applications, including indoor \cite{yang2017support}, video \cite{gay2019visual,mavroudi2020representation}, and 3D scene understanding \cite{wald2020learning}, as well as for providing scene explanations \cite{shi2019explainable}. Our approach also generates scene graphs, but uniquely employs qualitative calculus to formalize explanations for individual actions within the scene.

In automated driving, scene understanding must encompass vehicles, vulnerable road users, and stationary elements (e.g., traffic lights, cones, barriers), along with their movements. Scene graphs have been applied for overall scene understanding \cite{guo2012semantic1,greve2023collaborative}, lane estimation \cite{zurn2021lane}, and action prediction \cite{kochakarn2023explainable}. Our work showcases scene graphs for relevant interaction identification and extends to a domain-independent representation using knowledge graphs.

Dubba et al. \cite{dubba2015learning} propose an inductive logic programming framework to explain actions like aeroplane arrivals or departures through object relation chains from videos. This method, based on labelled examples, faces challenges such as manual data labelling and sensitivity to initial examples, often resulting in multiple predictions and false positives.

In \cite{zhuo2019explainable} and \cite{li2018effect}, actions are recognized by describing them as state transitions (e.g., an open microwave or a hand holding a cup). This involves training detectors for objects and states, using them to identify state transitions, and recognizing actions when these transitions match predefined descriptions. Despite high recall, this approach can produce many false positives.

Hua et al. \cite{hua2022towards} introduced a method for action recognition and explanation by representing videos with spatial algebra (RCC), dividing videos into clips, and extracting object relation chains. 
A neural saliency estimator scores these chains, with the highest-scored chains generating human-understandable explanations. 
This method, trained with a cross-entropy loss function, has proven effective in recognizing actions and providing clear explanations.
 
Recently, \cite{belmecheri2024trustworthy} proposed a method to build qualitative representations from BEV perception to generating explanations in automated driving, underscoring the importance of qualitative relations for explaining decision-making in autonomous systems.

Graph neural networks have been applied in other contexts of automated driving \cite{rahmani2023graph}, for example for pedestrian trajectory prediction \cite{luo2023gsgformer,diehl2019graph}, behaviour planning \cite{klimke2022cooperative}, but also for abnormal information identification in scene understanding \cite{jin2022graph,monninger2023scene,rong2024driving}, however without the use of the QXG as an intermediate representation.

\section{Method}

In the following, we describe how to apply GNNs for QXG explanations. Our method uses the predictive power of the model to classify the corresponding objects for an explanation, directly from the QXG representation.

\subsection{Relevant Object Identification}

Relevant object identification (ROI) is formulated as a binary edge classification problem.
Instead of directly classifying the relevance of each node in the graph, we classify the object relation chain as described by the edge between two nodes.
We choose this problem formulation, because (a) the edges carry most of the information in the graph, (b) it still aggregates the information from the surrounding nodes, and (c) it is closer to the edge list format used in previous work \cite{belmecheri2024trustworthy}.
Still, a formulation via direct node classification would be similarly possible.

\subsection{Qualitative Explainable Graph}

We describe the structure and features of the QXG used to describe the scene for the GNN.

We follow the general QXG setup as described in Section~\ref{sec:qxg} using qualitative distance calculi, qualitative trajectory calculi, and rectangle algebra (RA) as qualitative spatio-temporal relationships between objects.
Rectangle Algebra is encoded as two features (one for the x- and one for the y-axis), each other relation is encoded as a single edge feature.
For the nodes, the only feature is the object type, e.g. truck, pedestrian, ego vehicle.

\subsection{Graph Neural Network Architecture}

Graph neural networks are a specific paradigm of neural networks that can process graph-structured data~\cite{bacciu2020gentle}.
They process data iteratively by propagating information between nodes along the edges, followed by standard neural network layers to produce outputs per node, per edge, or globally for the entire graph, depending on the task.
Usually, the GNN layers are followed by one or multiple dense feed-forward layers to map the returned graph embeddings onto a real-valued vector for classification or a single output for regression tasks.

In the context of this paper, we apply graph attentional (GAT) layers ~\cite{DBLP:conf/iclr/VelickovicCCRLB18}.
GATs combine graph convolutional neural networks with the attention mechanism~\cite{DBLP:journals/corr/BahdanauCB14}, which is the key building block of the popular transformer architecture~\cite{vaswani2017attention}.

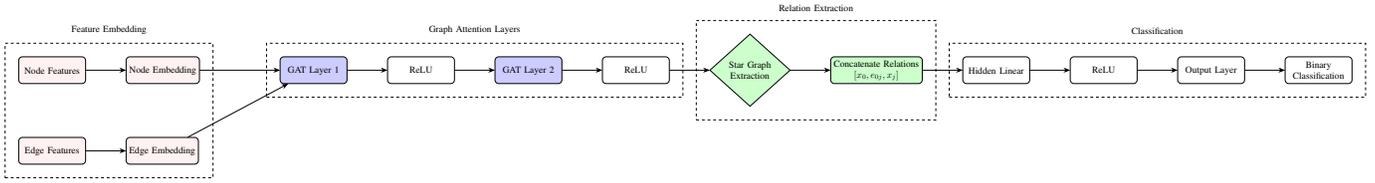
\begin{figure*}[t]
    \centering
    \resizebox{\textwidth}{!}{%
    \begin{tikzpicture}[
    box/.style={rectangle, draw, minimum width=2.5cm, minimum height=1cm, align=center, rounded corners},
    embedding/.style={box, fill=pink!20},
    attention/.style={box, fill=blue!20},
    extraction/.style={box, fill=green!20},
    decision/.style={diamond, draw, fill=green!20, align=center, minimum width=3cm, minimum height=2cm},
    arrow/.style={-{Stealth}, thick},
    node distance=1.5cm
]
    \node[embedding] (node_feat) {Node Features};
    \node[embedding] (edge_feat) [below=2cm of node_feat] {Edge Features};

    \node[embedding] (node_emb) [right=of node_feat] {Node Embedding};
    \node[embedding] (edge_emb) [right=of edge_feat] {Edge Embedding};

    \node[attention] (gat1) [right=3cm of node_emb] {GAT Layer 1};
    \node[box] (relu1) [right=of gat1] {ReLU};

    \node[attention] (gat2) [right=of relu1] {GAT Layer 2};
    \node[box] (relu2) [right=of gat2] {ReLU};

    \node[decision] (star) [right=of relu2] {Star Graph\\Extraction};

    \node[extraction] (concat) [right=of star] 
        {Concatenate Relations\\$[x_0, e_{0j}, x_j]$};

    \node[box] (hidden) [right=of concat] {Hidden Linear};
    \node[box] (relu3) [right=of hidden] {ReLU};
    \node[box] (output) [right=of relu3] {Output Layer};
    \node[box] (class) [right=of output] {Binary\\Classification};

    \draw[arrow] (node_feat) -- (node_emb);
    \draw[arrow] (edge_feat) -- (edge_emb);
    \draw[arrow] (node_emb) -- (gat1);
    \draw[arrow] (edge_emb) -- (gat1);
    \draw[arrow] (gat1) -- (relu1);
    \draw[arrow] (relu1) -- (gat2);
    \draw[arrow] (gat2) -- (relu2);
    \draw[arrow] (relu2) -- (star);
    \draw[arrow] (star) -- (concat);
    \draw[arrow] (concat) -- (hidden);
    \draw[arrow] (hidden) -- (relu3);
    \draw[arrow] (relu3) -- (output);
    \draw[arrow] (output) -- (class);

    \begin{scope}[on background layer]
        \node[fit=(node_feat)(edge_feat)(node_emb)(edge_emb),
              draw, dashed, inner sep=0.5cm] (emb_box) {};
        \node[above=0.2cm of emb_box] {Feature Embedding};
        
        \node[fit=(gat1)(relu1)(gat2)(relu2),
              draw, dashed, inner sep=0.5cm] (gat_box) {};
        \node[above=0.2cm of gat_box] {Graph Attention Layers};
        
        \node[fit=(star)(concat),
              draw, dashed, inner sep=0.5cm] (rel_box) {};
        \node[above=0.2cm of rel_box] {Relation Extraction};
        
        \node[fit=(hidden)(relu3)(output)(class),
              draw, dashed, inner sep=0.5cm] (class_box) {};
        \node[above=0.2cm of class_box] {Classification};
    \end{scope}
\end{tikzpicture}}
    \caption{Schematic Overview of the GNN Architecture}
    \label{fig:architecture}
\end{figure*}

Our architecture employs a multi-stage feature transformation and graph convolution approach and works as follows (see Figure~\ref{fig:architecture} for a visual overview):
First, we embed the single categorical node feature and the six edge features into a vector representation.
The edge features are first embedded individually and then the jointly aligned through an additional linear layer.

Then, the embedded graph representation is passed through two GAT layers, each with 4 attention heads, and ReLU activation functions in-between.

We then extract the representations of the star graph centred on the ego vehicle. 
The representation learning procedure involves concatenating node and edge features for each object pair, formally defined as:
$v_{ObjRelChain} = [x_0 \| e_{0j} \| x_j]$ where $x_0$ is the ego vehicle node representation, $e_{0j}$ are the embedded edge features connecting the ego vehicle to the target object and $x_j$ is the target object node representation. 
Together they form the embedded  object relation chain.
The joint representation vector is passed through another hidden linear layer, one ReLU activation, and finally the output layer with one output for binary classification.

\subsection{Model Training}

We train the model in a standard supervised manner with mini-batches of data over 100 epochs with the Adam optimizer (learning rate = $3 \times 10^{-4}$).
However, since the data is highly imbalanced with only one relevant object per frame in the training data, we must account for that in the training procedure.
To this end, we apply a combination of two loss functions that handle imbalanced data: 
Weighted Binary Cross-Entropy loss and focal loss \cite{DBLP:journals/pami/LinGGHD20}.

Weighted Binary Cross-Entropy loss (wBCE) is a modified version of the standard cross-entropy loss that accounts for the mismatch in number of positive and negative samples by weighing the corresponding loss term.
It is defined as
\begin{equation}
L_{wBCE} = -\frac{1}{N} \sum_{i=1}^{N} [w_p y_i log(\hat{y_i}) + w_n (1-y_i) log (1-\hat{y_i})]
\end{equation}
where $N$ is the mini-batch size, $w_p$ is the weight of positive examples, $w_n$ is the weight of negative examples, $y_i$ is the prediction score and $\hat{y_i}$ is the true label.

The focal loss (FL) \cite{DBLP:journals/pami/LinGGHD20} was introduced in the computer vision community for dense object recognition in images, a problem which is highly imbalanced, too.
It is defined as
\begin{equation}
    L_{FL} = -\frac{1}{N} \sum_{i=1}^{N} \alpha_t (1-p_t)^\gamma log(p_t)
\end{equation}
with \(p_t = y\hat{y} + (1-y)(1-\hat{y})\). The focal loss introduces two additional parameters: \(\gamma\) is a modulating factor to focus on easier or harder instances, and \(\alpha\) balances the importance of positive and negative examples. In our experiments, we set \(alpha=0.95\) and \(\gamma=0.5\) to put an emphasis on the few positive instances in the training data.

Our total loss is finally constructed as a weighted sum \(L = w * L_{wBCE} + L_{FL}\) with \(w=0.5\). We find that a single loss is not sufficient to learn a classifier in the imbalanced data regime for relevant object identification, as we will show in the experiments.

\section{Experiments}

\subsection{Experimental Setup}

\paragraph{Dataset}
Our evaluation is based on a real-world, large-scale dataset called nuScenes \cite{Caesar2020} that has been enriched with human labels of relevant objects in the additional DriveLM dataset \cite{sima2023drivelm}, as was done in previous work \cite{belmecheri2024roi}.
We take the subset of scenes and frames where at least one object was labelled as relevant by the DriveLM annotators. 
This leads to a total dataset of 2465 QXGs.

\paragraph{Baselines}
We consider two of the models proposed in \cite{belmecheri2024roi} as baselines, a random forest~\cite{breiman2001random} and AdaBoost~\cite{freund1997decision}.
We acknowledge that the selected baselines are traditional, shallow ML techniques; however, they are the most efficient models proposed in the existing literature on the QXG.
The baseline models are trained on the same dataset, but in a slightly different manner than the GNN.
Since they require a fixed input representation, they are not by themselves capable to handle varying graph sizes. To apply them, the ROI task is designed as a classification task where the edge list of the graphs is classified. 
That means, an object pair with the corresponding relations is classified as relevant or not, but without having the context of the other objects in the scene. The training is then performed over all object pairs in the dataset, independent of the scene and frame they appear in.
This differs from the GNN training, where each QXG represents a whole scene and is a single input to the model.

\paragraph{Metrics}
We measure the accuracy (number of correctly classified edges), F1 score (harmonized mean of precision and recall), precision (fraction of correct positive classification; \(\text{Precision} = \frac{\text{True Positives}}{\text{True Positives} + \text{False Positives}}\)), recall (fraction of correct negative classification; \(\text{Recall} = \frac{\text{True Positives}}{\text{True Positives} + \text{False Negatives}}\)), and ROC-AUC (Area Under the Receiver Operating Characteristic Curve).
We choose these metrics since they are established for the binary classification task, but also suitable to measure performance for classifiers under imbalanced data.
For all metrics, a larger value is better.

\paragraph{Technical Setup}
The GNN and its training are implemented in PyTorch Geometric 2.6.1 \cite{Fey/Lenssen/2019}, the baselines are implemented with scikit-learn 1.6.0 \cite{scikit-learn}.
All experiments are executed on a MacBook Pro 2023 with 32\,GB RAM.

\subsection{Results}

\subsubsection{Relevant Object Identification}

In our main case study, relevant object identification, we evaluate our GNN approach on the QXGs extracted from nuScenes with DriveLM relevance labels.
We perform 10-fold cross-validation, i.e., 1 part of the data is set aside for testing, 9 parts are used for training. This is repeated 10 times until each part of the data was used once for testing. All presented results are averaged over all 10 folds.
The results are shown in Table~\ref{tab:results}.

\begin{table}[t]
    \centering
    \caption{Results for relevant object identification.}
    \label{tab:results}
    \begin{tabular}{lrrrrr}
        \toprule
        Model & Accuracy & F1 & Precision & Recall & ROC-AUC\\\midrule
        GNN (ours) & 87.11 & 27.28 & 17.58 & 63.19 & 86.39 \\
        RF~\cite{breiman2001random} & 70.79 & 15.72 & 8.83 & 71.90 & 71.32 \\
        AdaBoost~\cite{freund1997decision} & 71.86 & 16.76 & 9.44 & 74.69 & 73.22 \\
        \bottomrule
    \end{tabular}
\end{table}

The experimental results demonstrate key findings about applying GNNs to QXGs for relevant object identification. First, our GNN-based approach shows substantial improvements over the baseline methods across all metrics, particularly in F1 and ROC-AUC scores. This suggests that considering the entire graph structure, rather than isolated relation chains, provides valuable context for identifying relevant objects in traffic scenes.

\subsubsection{Relevance of Loss Functions}

We perform an additional set of experiments to evaluate the selection of the combined loss function in the training. 
The results are shown in Table~\ref{tab:losses}.
Both wBCE and FL alone result in comparable models, even though with more imbalanced metrics; wBCE having stronger recall, whereas FL having stronger precision. Their combination leads to a well-rounded result that balances both advantages, even though at the cost of a few percentage points in accuracy/precision over FL alone.
Standard BCE alone does not lead to a competitive model, i.e. we see that is necessary to pick a loss function suitable for the imbalanced data regime of ROI.

\begin{table}[t]
    \centering
    \caption{Relevance of Loss Functions}
    \label{tab:losses}
    \begin{tabular}{lrrrrr}
        \toprule
        Model & Accuracy & F1 & Precision & Recall & ROC-AUC\\\midrule
        wBCE + FL & 87.11 & 27.28 & 17.58 & 63.19 & 86.39 \\
        wBCE only & 86.00 & 25.84 & 16.24 & 64.17 & 85.21 \\
        FL only & 89.51 & 28.54 & 19.40 & 54.85 & 85.31 \\
        BCE only & 96.22 & 11.66 & 52.18 & 6.66 & 85.70 \\
        \bottomrule
    \end{tabular}
\end{table}

\subsection{Limitations}

Some limitations of our approach should be noted. First, the model's performance heavily depends on the quality and completeness of the input QXG representation. 
Missing or incorrect object detections in the perception system would propagate through to the final predictions. 
Second, while our approach improves upon previous methods, the relatively low F1 scores (27.28\,\%) indicate that there is still room for improvement in ROI, which is potentially possible to overcome through additional training data and more extensive graph features.

From an explainability perspective, while QXGs provide an interpretable intermediate representation, the internal workings of the GNN layers remain somewhat opaque. Future work could explore methods to extract more detailed explanations from the learned attention weights and node embeddings.

Finally, our evaluation is limited to the nuScenes dataset with DriveLM annotations. While this provides a realistic test bed, the model's generalization to other datasets, sensor configurations, or driving scenarios remains to be validated. Future work should investigate the robustness of our approach across different environmental conditions and traffic scenarios.

\section{Conclusion}

This paper explores the combination of qualitative explainable graphs (QXG) and graph neural networks (GNN) for explainable scene understanding in automated driving, specifically relevant object identification.
While previous work demonstrated the effectiveness of QXGs using shallow machine learning models, our research advances the field by leveraging deep learning techniques that can process entire graph structures rather than isolated relation chains. 
The proposed GNN architecture, which incorporates graph attentional layers and a carefully designed loss function for imbalanced data, shows promising results in relevant object identification tasks.
Using GNNs is a flexible approach and can process scenes of varying size and density, making it compatible with a wide range of traffic scenarios. Here, the incorporation of the QXG is especially useful as a foundation to model the driving scenario in a meaningful representation.

Our experimental evaluation on the nuScenes dataset demonstrates that the GNN-based approach outperforms traditional machine learning baselines across multiple metrics. 
The model successfully balances the challenges of highly imbalanced data while maintaining the interpretability advantages of the QXG representation. 
This is particularly evident in the improved F1 score and ROC-AUC scores compared to random forest and AdaBoost baselines.

The results suggest that deep learning approaches can effectively complement qualitative spatial-temporal representations while maintaining explainability. 
In future work, we will explore extending this framework to other automated driving tasks, such as action explanation and prediction, under consideration of additional datasets, as well as investigating more sophisticated GNN architectures and attention mechanisms to further boost and understand the effectiveness of our approach.

\bibliographystyle{IEEEtran}
\bibliography{refs}

\begin{thebibliography}{10}
\providecommand{\url}[1]{#1}
\csname url@samestyle\endcsname
\providecommand{\newblock}{\relax}
\providecommand{\bibinfo}[2]{#2}
\providecommand{\BIBentrySTDinterwordspacing}{\spaceskip=0pt\relax}
\providecommand{\BIBentryALTinterwordstretchfactor}{4}
\providecommand{\BIBentryALTinterwordspacing}{\spaceskip=\fontdimen2\font plus
\BIBentryALTinterwordstretchfactor\fontdimen3\font minus
  \fontdimen4\font\relax}
\providecommand{\BIBforeignlanguage}[2]{{%
\expandafter\ifx\csname l@#1\endcsname\relax
\typeout{** WARNING: IEEEtran.bst: No hyphenation pattern has been}%
\typeout{** loaded for the language `#1'. Using the pattern for}%
\typeout{** the default language instead.}%
\else
\language=\csname l@#1\endcsname
\fi
#2}}
\providecommand{\BIBdecl}{\relax}
\BIBdecl

\bibitem{yang_scene_2019}
S.~Yang, W.~Wang, C.~Liu, and W.~Deng, ``Scene understanding in deep
  learning-based end-to-end controllers for autonomous vehicles,'' \emph{IEEE
  Transactions on Systems, Man, and Cybernetics: Systems}, vol.~49, no.~1, pp.
  53--63, 2019.

\bibitem{muhammad_vision-based_2022}
K.~Muhammad, T.~Hussain, H.~Ullah, J.~D. Ser, M.~Rezaei, N.~Kumar, M.~Hijji,
  P.~Bellavista, and V.~H.~C. de~Albuquerque, ``Vision-based semantic
  segmentation in scene understanding for autonomous driving: Recent
  achievements, challenges, and outlooks,'' \emph{{IEEE} Trans. Intell. Transp.
  Syst.}, vol.~23, no.~12, pp. 22\,694--22\,715, 2022.

\bibitem{muhammad_deep_2021}
K.~Muhammad, A.~Ullah, J.~Lloret, J.~D. Ser, and V.~H.~C. de~Albuquerque,
  ``Deep learning for safe autonomous driving: Current challenges and future
  directions,'' \emph{{IEEE} Trans. Intell. Transp. Syst.}, vol.~22, no.~7, pp.
  4316--4336, 2021.

\bibitem{belmecheri2023acquiring}
N.~Belmecheri, A.~Gotlieb, N.~Lazaar, and H.~Spieker, ``Acquiring qualitative
  explainable graphs for automated driving scene interpretation,'' \emph{CoRR},
  vol. abs/2308.12755, 2023.

\bibitem{belmecheri_2024_tra}
------, ``Trustworthy automated driving through qualitative scene understanding
  and explanations,'' in \emph{Transport Research Arena (TRA)}, 2024.

\bibitem{belmecheri2024trustworthy}
------, ``Toward trustworthy automated driving through qualitative scene
  understanding and explanations,'' \emph{SAE International Journal of
  Connected and Automated Vehicles}, vol.~8, 2024.

\bibitem{belmecheri2024roi}
------, ``Relevant object identification from qualitative explainable graphs in
  automated driving,'' \emph{NORA Annual Conference}, 2024.

\bibitem{renz_qualitative_2007}
J.~Renz and B.~Nebel, ``Qualitative {Spatial} {Reasoning} {Using} {Constraint}
  {Calculi},'' in \emph{Handbook of {Spatial} {Logics}}.\hskip 1em plus 0.5em
  minus 0.4em\relax Springer, 2007, pp. 161--215.

\bibitem{dylla_survey_2015}
F.~Dylla, J.~H. Lee, T.~Mossakowski, T.~Schneider, A.~van Delden, J.~van~de
  Ven, and D.~Wolter, ``A survey of qualitative spatial and temporal calculi:
  Algebraic and computational properties,'' \emph{{ACM} Comput. Surv.},
  vol.~50, no.~1, pp. 7:1--7:39, 2017.

\bibitem{yang2017support}
M.~Y. Yang, W.~Liao, H.~Ackermann, and B.~Rosenhahn, ``On support relations and
  semantic scene graphs,'' \emph{ISPRS journal of photogrammetry and remote
  sensing}, vol. 131, pp. 15--25, 2017.

\bibitem{gay2019visual}
P.~Gay, J.~Stuart, and A.~Del~Bue, ``Visual graphs from motion (vgfm): Scene
  understanding with object geometry reasoning,'' in \emph{Computer
  Vision--ACCV 2018: 14th Asian Conference on Computer Vision, Perth,
  Australia, December 2--6, 2018, Revised Selected Papers, Part III 14}.\hskip
  1em plus 0.5em minus 0.4em\relax Springer, 2019, pp. 330--346.

\bibitem{mavroudi2020representation}
E.~Mavroudi, B.~B. Haro, and R.~Vidal, ``Representation learning on
  visual-symbolic graphs for video understanding,'' in \emph{European
  Conference on Computer Vision}.\hskip 1em plus 0.5em minus 0.4em\relax
  Springer, 2020, pp. 71--90.

\bibitem{wald2020learning}
J.~Wald, H.~Dhamo, N.~Navab, and F.~Tombari, ``Learning 3d semantic scene
  graphs from 3d indoor reconstructions,'' in \emph{Proceedings of the IEEE/CVF
  Conference on Computer Vision and Pattern Recognition}, 2020, pp. 3961--3970.

\bibitem{shi2019explainable}
J.~Shi, H.~Zhang, and J.~Li, ``Explainable and explicit visual reasoning over
  scene graphs,'' in \emph{Proceedings of the IEEE/CVF conference on computer
  vision and pattern recognition}, 2019, pp. 8376--8384.

\bibitem{guo2012semantic1}
C.~Guo and S.~Mita, ``A semantic graph of traffic scenes for intelligent
  vehicle systems,'' \emph{IEEE intelligent systems}, pp. 57--62, 2012.

\bibitem{greve2023collaborative}
E.~Greve, M.~B{\"u}chner, N.~V{\"o}disch, W.~Burgard, and A.~Valada,
  ``Collaborative dynamic 3d scene graphs for automated driving,''
  \emph{arXiv:2309.06635}, 2023.

\bibitem{zurn2021lane}
J.~Z{\"u}rn, J.~Vertens, and W.~Burgard, ``Lane graph estimation for scene
  understanding in urban driving,'' \emph{IEEE Robotics and Automation
  Letters}, vol.~6, no.~4, pp. 8615--8622, 2021.

\bibitem{kochakarn2023explainable}
P.~Kochakarn, D.~De~Martini, D.~Omeiza, and L.~Kunze, ``Explainable action
  prediction through self-supervision on scene graphs,'' in \emph{2023 IEEE
  International Conference on Robotics and Automation (ICRA)}.\hskip 1em plus
  0.5em minus 0.4em\relax IEEE, 2023, pp. 1479--1485.

\bibitem{scene_graph_survey}
X.~Chang, P.~Ren, P.~Xu, Z.~Li, X.~Chen, and A.~Hauptmann, ``A comprehensive
  survey of scene graphs: Generation and application,'' \emph{IEEE Transactions
  on Pattern Analysis and Machine Intelligence}, vol.~45, no.~1, pp. 1--26,
  2023.

\bibitem{dubba2015learning}
K.~S. Dubba, A.~G. Cohn, D.~C. Hogg, M.~Bhatt, and F.~Dylla, ``Learning
  relational event models from video,'' \emph{Journal of Artificial
  Intelligence Research}, vol.~53, pp. 41--90, 2015.

\bibitem{zhuo2019explainable}
T.~Zhuo, Z.~Cheng, P.~Zhang, Y.~Wong, and M.~Kankanhalli, ``Explainable video
  action reasoning via prior knowledge and state transitions,'' in
  \emph{Proceedings of the 27th acm international conference on multimedia},
  2019, pp. 521--529.

\bibitem{li2018effect}
D.~Li, E.~Scala, P.~Haslum, and S.~Bogomolov, ``Effect-abstraction based
  relaxation for linear numeric planning.'' in \emph{IJCAI}, 2018, pp.
  4787--4793.

\bibitem{hua2022towards}
H.~Hua, D.~Li, R.~Li, P.~Zhang, J.~Renz, and A.~Cohn, ``Towards explainable
  action recognition by salient qualitative spatial object relation chains,''
  in \emph{Proceedings of the AAAI Conference on Artificial Intelligence
  (AAAI-22)}, 2022.

\bibitem{rahmani2023graph}
S.~Rahmani, A.~Baghbani, N.~Bouguila, and Z.~Patterson, ``Graph neural networks
  for intelligent transportation systems: A survey,'' \emph{IEEE Transactions
  on Intelligent Transportation Systems}, vol.~24, no.~8, pp. 8846--8885, 2023.

\bibitem{luo2023gsgformer}
Z.~Luo, M.~Robin, and P.~Vasishta, ``Gsgformer: Generative social graph
  transformer for multimodal pedestrian trajectory prediction,'' \emph{arXiv
  preprint arXiv:2312.04479}, 2023.

\bibitem{diehl2019graph}
F.~Diehl, T.~Brunner, M.~T. Le, and A.~Knoll, ``Graph neural networks for
  modelling traffic participant interaction,'' in \emph{2019 IEEE Intelligent
  Vehicles Symposium (IV)}.\hskip 1em plus 0.5em minus 0.4em\relax IEEE, 2019,
  pp. 695--701.

\bibitem{klimke2022cooperative}
M.~Klimke, B.~V{\"o}lz, and M.~Buchholz, ``Cooperative behavior planning for
  automated driving using graph neural networks,'' in \emph{2022 IEEE
  Intelligent Vehicles Symposium (IV)}.\hskip 1em plus 0.5em minus 0.4em\relax
  IEEE, 2022, pp. 167--174.

\bibitem{jin2022graph}
K.~Jin, H.~Wang, C.~Liu, Y.~Zhai, and L.~Tang, ``Graph neural network based
  relation learning for abnormal perception information detection in
  self-driving scenarios,'' in \emph{2022 International Conference on Robotics
  and Automation (ICRA)}.\hskip 1em plus 0.5em minus 0.4em\relax IEEE, 2022,
  pp. 8943--8949.

\bibitem{monninger2023scene}
T.~Monninger, J.~Schmidt, J.~Rupprecht, D.~Raba, J.~Jordan, D.~Frank, S.~Staab,
  and K.~Dietmayer, ``Scene: Reasoning about traffic scenes using heterogeneous
  graph neural networks,'' \emph{IEEE Robotics and Automation Letters}, vol.~8,
  no.~3, pp. 1531--1538, 2023.

\bibitem{rong2024driving}
F.~Rong, W.~Peng, M.~Lan, Q.~Zhang, and L.~Zhang, ``Driving scene understanding
  with traffic scene-assisted topology graph transformer,'' in
  \emph{Proceedings of the 32nd ACM International Conference on Multimedia},
  2024, pp. 10\,075--10\,084.

\bibitem{bacciu2020gentle}
D.~Bacciu, F.~Errica, A.~Micheli, and M.~Podda, ``A gentle introduction to deep
  learning for graphs,'' \emph{Neural Networks}, vol. 129, pp. 203--221, 2020.

\bibitem{DBLP:conf/iclr/VelickovicCCRLB18}
\BIBentryALTinterwordspacing
P.~Velickovic, G.~Cucurull, A.~Casanova, A.~Romero, P.~Li{\`{o}}, and
  Y.~Bengio, ``Graph attention networks,'' in \emph{6th International
  Conference on Learning Representations, {ICLR}}.\hskip 1em plus 0.5em minus
  0.4em\relax OpenReview.net, 2018. [Online]. Available:
  \url{https://openreview.net/forum?id=rJXMpikCZ}
\BIBentrySTDinterwordspacing

\bibitem{DBLP:journals/corr/BahdanauCB14}
\BIBentryALTinterwordspacing
D.~Bahdanau, K.~Cho, and Y.~Bengio, ``Neural machine translation by jointly
  learning to align and translate,'' in \emph{3rd International Conference on
  Learning Representations, {ICLR} 2015, San Diego, CA, USA, May 7-9, 2015,
  Conference Track Proceedings}, Y.~Bengio and Y.~LeCun, Eds., 2015. [Online].
  Available: \url{http://arxiv.org/abs/1409.0473}
\BIBentrySTDinterwordspacing

\bibitem{vaswani2017attention}
A.~Vaswani, ``Attention is all you need,'' \emph{Advances in Neural Information
  Processing Systems}, 2017.

\bibitem{DBLP:journals/pami/LinGGHD20}
\BIBentryALTinterwordspacing
T.~Lin, P.~Goyal, R.~B. Girshick, K.~He, and P.~Doll{\'{a}}r, ``Focal loss for
  dense object detection,'' \emph{{IEEE} Trans. Pattern Anal. Mach. Intell.},
  vol.~42, no.~2, pp. 318--327, 2020. [Online]. Available:
  \url{https://doi.org/10.1109/TPAMI.2018.2858826}
\BIBentrySTDinterwordspacing

\bibitem{Caesar2020}
H.~Caesar, V.~Bankiti, A.~H. Lang, S.~Vora, V.~E. Liong, Q.~Xu, A.~Krishnan,
  Y.~Pan, G.~Baldan, and O.~Beijbom, ``nuscenes: {A} multimodal dataset for
  autonomous driving,'' in \emph{2020 {IEEE/CVF} Conference on Computer Vision
  and Pattern Recognition, {CVPR} 2020, Seattle, WA, USA, June 13-19,
  2020}.\hskip 1em plus 0.5em minus 0.4em\relax Computer Vision Foundation /
  {IEEE}, 2020, pp. 11\,618--11\,628.

\bibitem{sima2023drivelm}
C.~Sima, K.~Renz, K.~Chitta, L.~Chen, H.~Zhang, C.~Xie, P.~Luo, A.~Geiger, and
  H.~Li, ``Drivelm: Driving with graph visual question answering,''
  \emph{CoRR}, vol. abs/2312.14150, 2023.

\bibitem{breiman2001random}
L.~Breiman, ``Random forests,'' \emph{Machine learning}, vol.~45, no.~1, pp.
  5--32, 2001.

\bibitem{freund1997decision}
Y.~Freund and R.~E. Schapire, ``A decision-theoretic generalization of on-line
  learning and an application to boosting,'' \emph{Journal of computer and
  system sciences}, vol.~55, no.~1, pp. 119--139, 1997.

\bibitem{Fey/Lenssen/2019}
M.~Fey and J.~E. Lenssen, ``Fast graph representation learning with {PyTorch
  Geometric},'' in \emph{ICLR Workshop on Representation Learning on Graphs and
  Manifolds}, 2019.

\bibitem{scikit-learn}
F.~Pedregosa, G.~Varoquaux, A.~Gramfort, V.~Michel, B.~Thirion, O.~Grisel,
  M.~Blondel, P.~Prettenhofer, R.~Weiss, V.~Dubourg, J.~VanderPlas, A.~Passos,
  D.~Cournapeau, M.~Brucher, M.~Perrot, and E.~Duchesnay, ``Scikit-learn:
  Machine learning in python,'' \emph{J. Mach. Learn. Res.}, vol.~12, pp.
  2825--2830, 2011.

\end{thebibliography}

\end{document}